\documentclass{article}

\usepackage{PRIMEarxiv}
\usepackage{xcolor}
\usepackage[utf8]{inputenc} 
\usepackage[T1]{fontenc}    
\usepackage{hyperref}       
\usepackage{url}            
\usepackage{booktabs}       
\usepackage{amsfonts}       
\usepackage{nicefrac}       
\usepackage{microtype}      
\usepackage{lipsum}
\usepackage{fancyhdr}       
\usepackage{graphicx}       
\graphicspath{{media/}}     

\title{Face Detection: Present State and Research Directions
\thanks{\textit{\underline{Citation}}: 
\textbf{Purnendu Prabhat, Himanshu Gupta and Ajeet Kumar Vishwakarma. Face Detection: Present State and Research Directions}} 
}

\author{
  Purnendu Prabhat \\
  Algoristan Research, \\
  Gurugram, Haryana, India  \\
  \texttt{mukul4world@gmail.com }\\
   \And
  Himanshu Gupta \\
  Tula’s Institute, \\
  Dehradun, Uttarakhand, India\\
  \texttt{himanshu.code11@gmail}\\
  (Corresponding author)\\
  \And
  Ajeet Kumar Vishwakarma \\
  Dev Bhoomi Uttarakhand University, \\
  Dehradun, Uttarakhand, India\\
  \texttt{ajeet7488@gmail} \\
}

\begin{document}
\maketitle

\begin{abstract}
\textbf{The majority of computer vision applications that handle images featuring humans use face detection as a core component. Face detection still has issues, despite much research on the topic. Face detection's accuracy and speed might yet be increased. This review paper shows the progress made in this area as well as the substantial issues that still need to be tackled. The paper provides research directions that can be taken up as research projects in the field of face detection.}
\end{abstract}

\keywords{Face detection \and Computer vision \and Artificial intelligence}

\section{Introduction}
Locating human faces in digital photos is done using a face detection algorithm, which is based on artificial intelligence (AI). Facial detection technology makes real-time surveillance and tracing of persons feasible in a variety of fields, including security, biometrics, law enforcement, entertainment, and personal security [1]. Face detection has progressed from traditional computer vision methods to more sophisticated artificial neural networks (ANNs) and associated technologies, with the ultimate result being a steady improvement in performance. It is the foundation for several important applications, including face tracking, face analysis, and face recognition.
Face detection helps with facial analysis by helping to select which areas of an image or video to focus on to identify age, gender and emotion from facial expressions. Face detection data is necessary for algorithms that determine which elements of an image or video are necessary to create a facial print in a facial recognition system that mathematically maps an individual's facial features and stores the data as a face print. If a new facial print is discovered, it can be compared to facial prints that have already been stored to see if there is a match. Figure 1 shows different face detection techniques.

Major contributions of this paper are a brief, yet comprehensive review of the approaches and advances in the field of face detection; and a list of challenges and research directions.

Section 2 discusses the applications of face detection, section 3 summarizes approaches for face detection while section 4 elucidates the challenges and research directions. Finally, section 5 concludes the paper.

\begin{figure}
    \centering
    \includegraphics[width=0.9\linewidth]{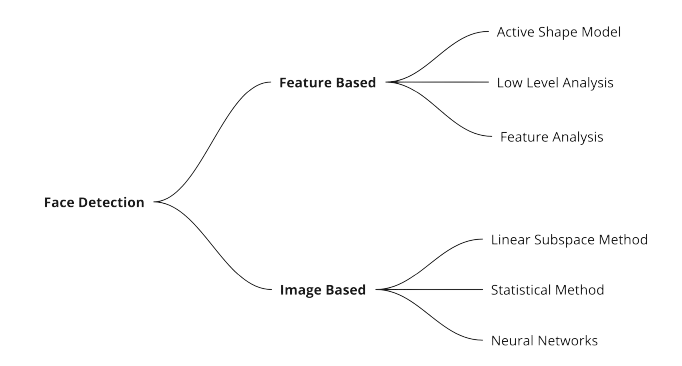}
    \caption{Face detection techniques.}
    \label{fig:enter-label}
\end{figure}

\section{Application of Face Detection}
\label{sec:headings}

In biometrics, face detection is often used in conjunction with or in addition to facial recognition systems. In addition, it is used in image database management, human-computer interaction, and video surveillance. Face tracking, face analysis, and face recognition are just some of the applications that start with face detection [2]. Facial analysis algorithms can identify age, gender, and emotion based on facial expressions, using face detection to tell the algorithms which areas of an image (or video) to focus on. In order for algorithms to create facial prints that are compared to previously recorded facial prints to see if there is a match, face detection is required in facial recognition [2]. Further, the use of Face Detection techniques can be seen in the following [3, 4]:
\begin{itemize}
    \item  Human computer interaction system
    \item  Photography: Face detection for auto focus and selecting regions of interest
    \item  Detecting focus and attentiveness of individuals
    \item  Counting number of people and crowd detection systems.
\end{itemize}

\section{Approaches for Face Detection}
\subsection{Featured based approach}
Features based approaches look for faces' invariant traits to aid in detection. The fundamental presumption is based on the fact that people can recognise faces and things with ease in a variety of positions and lighting situations, indicating that there must be traits or attributes that are invariant across these variations. Edge detectors are frequently used to extract facial elements such as skin tone, hairline, eyebrows, eyes, and other facial features. A statistical model is created based on the characteristics that were retrieved to characterise their correlations and confirm the presence of a face. Figure 2 shows different feature based techniques for face detection.

\subsubsection{Active shape model}

{Snakes[5-6], also known as active contour models, act like digital rubber bands, flexing and stretching to trace outlines such as faces in images. They analyze edges and shapes, adjusting to fit features like eyes, nose, and mouth, crucial for tasks like straightening faces in photos or interpreting facial expressions in apps. Alongside, active contour models, including Snakes, are vital for extracting facial characteristics.The Point Distribution Model (PDM), which was developed independently of computerised image analysis, was used to generate statistical models of form [7]. The idea is that forms may be statistically analysed using the same techniques as other multivariate objects if they can be represented as vectors. These models, often referred to as Point Distribution Models, create models by using principal components to learn the permitted constellations of form points from training instances. Deformable templates take into account a priori of facial features to improve the performance of Snakes[8]. Finding a facial feature boundary is a challenging undertaking since it is difficult to arrange the local evidence of face margins using generic contours into a coherent overall entity. The edge recognition procedure is complicated by some of these features’ low brightness contrast. By combining global information from the eye, the concept of snakes is taken a step further, enhancing the accuracy of the extraction procedure. The narrow valley, edge, peak, and brightness are the basis for deformation. The extraction of prominent features, such as the eyes, nose, mouth, and brows, is a major difficulty in face identification in addition to the face border.
\subsubsection{Low level analysis}

Various low-level analysis techniques for detecting facial features rely on different color models, including skin color-based approaches, RGB, HSV, YCbCr, and CIELAB. Skin color-based methods use predefined thresholds in RGB space to identify skin pixels, while RGB models represent colors using red, green, and blue channels. HSV separates color information into hue, saturation, and value components, useful for identifying skin tones. YCbCr separates luminance from chrominance, with specific values aiding in skin detection. CIELAB, though less common, offers perceptually uniform color space, enhancing color accuracy. Each model has advantages and challenges, and their selection depends on factors like lighting conditions and application requirements.

\subsubsection{Feature analysis}
These algorithms seek to discover faces by identifying structural traits that remain constant regardless of changes in stance, perspective, or illumination. These techniques are primarily intended for face localization. Viola and Jones, a famous name in the field of face detection, suggested a rapid and reliable face detection system that is 95 percent accurate and 15 times faster than previously used techniques[9]. The method is based on the usage of basic Haar-like characteristics that are swiftly evaluated through a new picture representation. To characterise the characteristics
of the picture texture, the Local Binary Pattern (LBP) approach is particularly useful[10]. Due to its benefits in terms of high-speed calculation and rotation invariance, LBP is widely used in areas such as image retrieval, texture analysis, face recognition, picture segmentation, etc.. Each pixel in LBP has a texture value attached to it that may be used as a target for tracking thermographic and monochromatic video. The target region's important features are identified using major uniform LBP patterns, which are then utilised to create a mask for choosing joint colour-texture features. AdaBoost creates the strong learner (a classifier that is well-correlated to the true classifier) by iteratively adding weak learners (a classifier that is only slightly correlated to the true classifier). There is no need for any prior knowledge of the face structure. When compared to other approaches, LBP-AdaBoost and Haar-AdaBoost are the quickest [11]. The Haar-AdaBoost technique, however, continues to be the best of the four methods in terms of detection rate and false detection rate. Other methods depend on Gabor and Gaussian filters[12,13].

\begin{figure}
    \centering
    \includegraphics[width=0.9\linewidth]{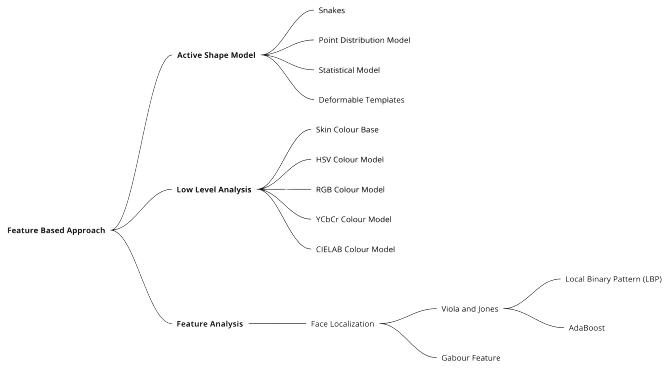}
    \caption{Feature based approaches for face detection.}
    \label{fig:enter-label}
\end{figure}
\subsection{Image based approach}
Images serve as examples for learning templates. The majority of appearance-based methods use statistical analysis and machine learning techniques to identify the pertinent aspects of face and non-facial photos. The discriminant functions or distribution models that result from the learnt attributes are then employed for face detection. Figure 3 shows different image based approaches for face detection. Major approaches in this domain include:
\subsubsection{Linear subspace method}

Eigenvectors are like special directions that help in understanding faces in pictures. With techniques like Principal Component Analysis (PCA), we can find these special directions, called eigenfaces, from a bunch of face photos that are made to look similar and organized. By using these eigenfaces, we can teach a simple computer program to recognize faces. It's fascinating because we can use just a few basic face images to teach the program, and it still does a good job at recognizing different faces. Moreover, a small number of basis images can be used to linearly encode facial image data[14], making the process computationally efficient and effective for face recognition tasks.
\begin{figure}
    \centering
    \includegraphics[width=0.9\linewidth]{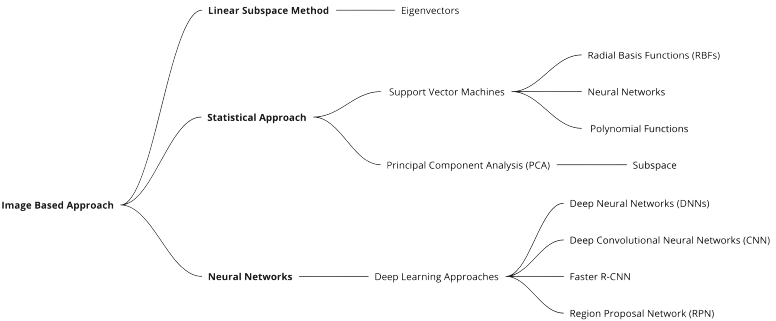}
    \caption{Image based approaches for face detection.}
    \label{fig:enter-label}
\end{figure}
\subsubsection{Statistical approach}
Face detection has been accomplished using Support Vector Machines (SVMs)[15]. SVMs operate as a novel paradigm for training classifiers based on radial basis functions (RBFs), neural networks, and polynomial functions. Principal Component Analysis (PCA) is also used for face detection. Onto the subspace, faces are projected and grouped. The same subspace is projected and clustered for non-face training images. The distance between each pixel of the image and the face space is calculated in order to identify faces in scenes. A face map is produced after computing the separation from face space.
\subsubsection{Neural Networks}
The benefit of using Deep Neural Networks (DNNs) for detection of images relieves us from finding the features beforehand. However, the dataset used to train such DNNs must be sufficiently large datasets, i.e., big data. This has been possible after the development of Graphical Processing Units (GPUs) and large scale distributed computing systems that can process big data. Detecting faces in a given image is an object detection task of computer vision. In recent years, major breakthroughs in face detection have been made utilising deep learning approaches, particularly deep Convolutional Neural Networks (CNN), which have shown exceptional success in a variety of computer vision applications. Deep learning methods, as opposed to traditional computer vision algorithms, skip the hand-crafted design process and have dominated numerous well-known benchmark assessments, such as the ImageNet Large Scale Visual Recognition Challenge (ILSVRC)[16]. Researchers recently employed the Faster R-CNN, one of the most advanced generic object detectors, and obtained encouraging results. Furthermore, end-to-end optimization was achieved by cooperative training on CNN cascade, Region Proposal Network (RPN), and faster R-CNN[17]. With hard negative mining and ResNet, the faster R-CNN face detection algorithm demonstrated considerable improvements in detection performance on face identification benchmarks such as FDDB [18,19,20].
\section{Challenges and Research Directions}
There has been good progress in the field of face detection.
However, the nature of computer vision applications are such that there has to be the next challenge available. Following are major challenges that can be next research direction in the field:
\begin{itemize}
    \item False Positives: Faces are detected even when there is no face. There can be following reasons for false positives in face detection:
        \begin{itemize}
            \item Fake faces: Printed images are also classified as faces of alive people.
            \item Poor training of the AI-model results in high false positives. Authors in [21] have designed a modified Adaboost algorithm to reduce false positives in face detection.
        \end{itemize}
    \item False Negatives: Faces are not detected even when there is a face. There can be following reasons for false positives in face detection:
        \begin{itemize}
            \item Masked Faces: Recent pandemic had us wear masks which was a major challenge in detecting faces. Also, spectacles and beards can sometimes be a hindrance in detecting faces.
            \item Poor lighting conditions are also an impediment in detecting faces in an image.
            \item Poor training of the AI-model
            \item Pose: Detecting faces from the side pose is difficult which worsens when there is bad lighting conditions.
        \end{itemize}
    \item Ethical concerns: There have been multiple instances of AI being racist and an infringement of individual privacy.
    \item Computational limitations: 
        \begin{itemize}
            \item Real-time performance on the Edge: We have been able to achieve real-time performance of face detection through Viola-Jones method[22]. However, the RAM and CPU requirements are the cost that we need to pay for real-time detection. An edge device that is ingesting feeds from multiple sources has limited RAM and CPU resources. This causes a bottleneck while processing multiple video streams in parallel. If real-time face detection on multiple video streams is optimised, then that can help in frame dropping as suggested by Ankur and Purnendu in [3].
        \end{itemize}
\end{itemize}

\section{Conclusion}
The paper discusses major approaches for face detection and provides a brief overview of the state of the art in this domain. There is still a need to go deeper, especially for real-time face detection on edge devices. The paper gives research directions that can be carried out in this domain.

\section*{References}

[1] https://www.techtarget.com/searchenterpriseai/definition/face-detection (Accessed: 03 Feb 2024 )

[2] https://sightcorp.com/knowledge-base/face-detection/ (Accessed: 03 Feb 2024)

[3] Gupta, Ankur, and Purnendu Prabhat. "Towards a resource efficient and privacy-preserving framework for campus-wide video analytics-based applications." Complex \& Intelligent Systems (2022): 1-16.

[4] Kumar, Ashu, Amandeep Kaur, and Munish Kumar. "Face detection techniques: a review." Artificial Intelligence Review 52.2 (2019): 927-948.

[5] Kass, Michael, Andrew Witkin, and Demetri Terzopoulos. "Snakes: Active contour models." International journal of computer vision 1.4 (1988): 321-331.

[6] Seo, Kap-Ho Seo Kap-Ho, et al. "Face detection and facial feature extraction using color snake." Industrial Electronics, 2002. ISIE 2002. Proceedings of the 2002 IEEE International Symposium On. Vol. 2. IEEE, 2002.

[7] Hjelmås, Erik, and Boon Kee Low. "Face detection: A survey." Computer vision and image understanding 83.3 (2001): 236-274.

[8] Yuille, Alan L., Peter W. Hallinan, and David S. Cohen. "Feature extraction from faces using deformable templates." International journal of computer vision 8.2 (1992): 99-111.

[9] Viola, Paul, and Michael J. Jones. "Robust real-time face detection." International journal of computer vision 57.2 (2004): 137-154.

[10] Ahonen, Timo, Abdenour Hadid, and Matti Pietikäinen. "Face recognition with local binary patterns." European conference on computer vision. Springer, Berlin, Heidelberg, 2004.

[11] Filali, Hajar, et al. "Multiple face detection based on machine learning." 2018 International Conference on Intelligent Systems and Computer Vision (ISCV). IEEE, 2018.
[12] Sharif, Muhammad, et al. "Face Recognition using Gabor Filters." Journal of Applied Computer Science and Mathematics 11 (2011).

[13] Young, Ian T., and Lucas J. Van Vliet. "Recursive implementation of the Gaussian filter." Signal processing 44.2 (1995): 139-151.

[14] Kirby, Michael, and Lawrence Sirovich. "Application of the Karhunen-Loeve procedure for the characterization of human faces." IEEE Transactions on Pattern analysis and Machine intelligence 12.1 (1990): 103-108.

[15] Mingxing, Jia, et al. "An improved detection algorithm of face with combining AdaBoost and SVM." 2013 25th Chinese Control and Decision Conference (CCDC). IEEE, 2013.

[16] Russakovsky, Olga, et al. "Imagenet large scale visual recognition challenge." International journal of computer vision 115.3 (2015): 211-252.

[17] Jiang, Huaizu, and Erik Learned-Miller. "Face detection with the faster R-CNN." 2017 12th IEEE international conference on automatic face \& gesture recognition (FG 2017). IEEE, 2017.

[18] He, Kaiming, et al. "Deep residual learning for image recognition." Proceedings of the IEEE conference on computer vision and pattern recognition. 2016.

[19] Lin, Kaihan, et al. "Face detection and segmentation based on improved mask R-CNN." Discrete dynamics in nature and society 2020 (2020).

[20] Jain, Vidit, and Erik Learned-Miller. Fddb: A benchmark for face detection in unconstrained settings. Vol. 2. No. 6. UMass Amherst technical report, 2010.

[21] Niyomugabo, Cesar, Hyo-rim Choi, and Tae Yong Kim. "A modified Adaboost algorithm to reduce false positives in face detection." Mathematical Problems in Engineering 2016 (2016).

[22] Savadatti, Mamta B., et al. "Theoretical Analysis of Viola-Jones Algorithm Based Image and Live-Feed Facial Recognition." 2022 International Conference on Advances in Computing, Communication and Applied Informatics (ACCAI). IEEE, 2022.

\end{document}